\pdfoutput=1
\documentclass[runningheads]{llncs}
\usepackage[utf8]{inputenc} 

\usepackage{textcomp}
\usepackage{graphicx}
\usepackage{comment}
\usepackage[dvipsnames]{xcolor}
\usepackage{array}
\usepackage{amsmath}
\usepackage{amsfonts}
\usepackage{caption}
\usepackage{dirtytalk}
\usepackage{hyperref}
\usepackage{cleveref}

\begin{document}
    \setlength{\parskip}{0pt}
\title{Weakly Supervised Semantic Segmentation Using Constrained Dominant Sets \thanks{This work is supported by The Scientific and Technological Research
Council of Turkey under TUBITAK BIDEB-2219 grant no 1059B191701102.}
\thanks{\textcolor{Purple}{Aslan S., Pelillo M. (2019) Weakly Supervised Semantic Segmentation Using Constrained Dominant Sets. In: Ricci E., Rota Bulò S., Snoek C., Lanz O., Messelodi S., Sebe N. (eds) Image Analysis and Processing (ICIAP 2019). Lecture Notes in Computer Science, vol 11752. Springer, Cham. \textbf{\textit{The final authenticated version is
available online at} \url{https://doi.org/10.1007/978-3-030-30645-8_39}}}}}
\titlerunning{Weakly Supervised Semantic Segmentation Using CDS's}
%
%
\author{Sinem Aslan\inst{1,2,3} \and
Marcello Pelillo\inst{1,2}}
\authorrunning{S. Aslan and M. Pelillo}
%
\institute{DAIS, Ca\textquotesingle \hspace{0.01cm} Foscari University of Venice, Italy \and
ECLT, Ca\textquotesingle \hspace{0.01cm} Foscari University of Venice, Italy \and
International Computer Institute, Ege University, Izmir, Turkey\\
\email{sinem.aslan@unive.it, pelillo@unive.it}}
\maketitle              
\begin{abstract}
The availability of large-scale data sets is an essential prerequisite for deep learning based semantic segmentation schemes. Since obtaining pixel-level labels is extremely expensive, supervising deep semantic segmentation networks using low-cost weak annotations has been an attractive research problem in recent years. In this work, we explore the potential of \emph{Constrained Dominant Sets} (\emph{CDS}) for generating multi-labeled full mask predictions to train a fully convolutional network (FCN) for semantic segmentation. Our experimental results show that using CDS's yields higher-quality mask predictions compared to methods that have been adopted in the literature for the same purpose.

\end{abstract}
\section{Introduction}
\vspace{-0.2cm}
\label{sec:int}
Semantic segmentation is one of the most well-studied research problems in computer vision. The goal is to achieve pixel-level classification, i.e., to label each pixel in a given input image with the class of the object or region that covers it. Predicting the class of each pixel yields to complete scene understanding which is the main problem of a wide range of computer vision applications, e.g. autonomous driving \cite{cordts2016cityscapes}, human-computer interaction \cite{oberweger2015hands}, earth observation \cite{audebert2016semantic}, biomedical applications \cite{wang2016deep}, dietary assessment systems \cite{Cioc1809Semantic}, etc. 
Stunning performances of DCNNs (Deep Convolutional Neural Networks) at image classification tasks have encouraged researchers to employ them for pixel-level classification as well. Outstanding methods in well-known benchmarks, e.g. PASCAL VOC 2012, train some fully convolutional networks (FCN) with supervision of fully-annotated ground-truth masks. However, obtaining such precise fully-annotated masks is extremely expensive and this limits the availability of large-scale annotated training sets for deep learning architectures. In order to address the aforementioned issue, recent works explored supervision of DCNN architectures for semantic segmentation using low-cost annotations like image-level labels \cite{huang2018weakly},
point tags \cite{bearman2016s},
bounding box \cite{khoreva2017simple,dai2015boxsup,Papandreou_2015_ICCV}  and scribbles\cite{lin2016scribblesup,vernaza2017learning,tang2018normalized,tang2018regularized}, that are weaker than the pixel-level labels.

Creating weak annotations is much easier than creating full annotations which 
helps to obtain large training sets for semantic segmentation. 
However, these annotations are not as precise as full annotations and their quality depends on the decisions made by the users, which degrades their reliability. Hence, literature works proposed different strategies for weakly-supervised semantic segmentation to deal with these issues. While a number of works \cite{tang2018normalized,tang2018regularized} 
proposed to employ a genuine cost function to get into account only the initially given true weak annotations at the training stage, another and the most common approach \cite{dai2015boxsup,lin2016scribblesup,khoreva2017simple,Papandreou_2015_ICCV,vernaza2017learning} has been supervising DCNN architectures by \emph{predicted full mask annotations} which are obtained by post-processing the weak-annotations. 

Among these two strategies, we follow the second one and propose to generate full mask annotations from scribbles by an interactive segmentation technique which has proven to be extremely effective in a variety of computer vision problems including image and video segmentation \cite{mequanint2018dominant,bulo2017dominant}
. For the same purpose, literature works have used a number of shallow interactive segmentation methods, e.g. variants of GrabCut \cite{rother2004grabcut} are used in \cite{Papandreou_2015_ICCV,khoreva2017simple} for propagating bounding box annotations to supervise a convolutional network. In order to propagate bounding box annotations, \cite{dai2015boxsup} proposed to perform iterative optimization between generating full mask approximations and training the network. 
Using a similar iterative scheme, \cite{lin2016scribblesup} propagated scribble annotations by superpixels via optimizing a multi-label graph cuts model of \cite{boykov2004experimental}. \cite{vernaza2017learning} proposed a random-walk based label propagation mechanism to propagate scribble annotations. 

In this paper, we aim to explore the potential of \emph{Constrained Dominant Sets} (\emph{CDS}) \cite{zemene2016interactive,mequanint2018dominant,LeulShahPel19,LeulPel18,ZemeneAP16b} for generating predicted full annotations to be used in supervision of a convolutional neural network for semantic segmentation. Representing images in an edge-weighted graph structure, main idea in constrained segmentation approach in 
\cite{mequanint2018dominant} is finding the collection of dominant set clusters on the graph that are constrained to contain the components of a given annotation. CDS approach is applied for co-segmentation and interactive segmentation using modalities of bounding box or scribble and superiority of it over the state of the art segmentation techniques like Graph Cut, Lazy Snapping, Geodesic Segmentation, Random Walker, Transduction, Geodesic Graph Cut, Constrained Random Walker is proved in \cite{mequanint2018dominant}. Motivated by the reported performance achievements for single cluster extraction (i.e. foreground extraction) in \cite{mequanint2018dominant}, we used CDS for multiple cluster extraction involving multi-label scribbles for the PASCAL VOC 2012 dataset. Since our goal is mainly exploring the performance of CDS in full mask prediction for weakly-supervised semantic segmentation, we trained a basic segmentation network, namely Fully Convolutional Network (FCN-8s) of \cite{long2015fully} based on VGG16 architecture, 
and compared our performance with other full mask prediction schemes in the literature that supervise the same type of deep learning architecture. 
Our experimental results on the standard dataset PASCAL VOC 2012 show the effectiveness of our approach compared to existing algorithms.

\section{Constrained Dominant Sets}
\vspace{-0.2cm}
\paragraph{Dominant Set Framework.} In the dominant-set clustering framework
\label{sec:cds} \cite{pavan2007dominant,bulo2017dominant}, an input image is represented as an undirected edge-weighted graph with no self-loops $G=(V,E,w)$, where $V=\{1,...,n\}$ is the set of vertices that correspond to image points (pixels or superpixels), $E \subseteq V \times V$ is the set of edges that represent the neighborhood relations between vertices, and $w=E \to R_+^*$ is the (positive) weight function that represent the similarity between linked node pairs. A symmetric affinity (or similarity) matrix is constructed to represent the graph $G$ that is denoted by $A=(a_{ij})_{n\times n}$ where $a_{ij}=w(i,j)$, if $(i,j)\in E$ and $a_{ij}=0$ otherwise.

Next, a weight $w_S(i)$, which is (recursively) defined as Eq. \ref{eq:over_sim}, is assigned to each vertex $i\in S$,
\vspace{-0.2cm}
\begin{equation}
\label{eq:over_sim}
    w_S(i) =     
    \begin{cases}
    1 
    & \quad \text{if } |S|=1, \\
    \sum_{j\in{S\setminus\{i\}}} \phi_{S\setminus\{i\}}(j,i) w_{S\setminus\{i\}}(j), & \quad \text{otherwise.}
    \end{cases}
\end{equation} 
where $\phi_S(i,j)$ denotes the (relative) similarity between nodes $j$ ($j \notin S$) and $i$, with respect to the average similarity between node $i$ and its neighbours in $S$ (defined by $\phi_S(i,j)=a_{ij}-\frac{1}{|S|}\sum_{k \in S}a_{ik}$). 

A positive $w_S(i)$ indicates that adding $i$ into its neighbours in $S$ will increase the internal coherence of the set, while when it is negative overall coherence gets decreased. Based on aforementioned definitions, a non-empty subset of vertices $S \subseteq V$ such that $\sum_{i \in T} w_T(i)>0$ for any non-empty $T \subseteq S$, is said to be \textit{dominant set} if it is a maximally coherent data set, i.e. satisfying two basic properties of a cluster that are \textit{internal coherence} ($w_S(i)>0$, for all $i \in S$) and \textit{external incoherence} ($w_{S\cap\{i\}} < 0$, for all $i \notin S$). 

Consider the following linearly-constrained quadratic optimization problem,
\vspace{-0.2cm}
\begin{equation}
\begin{aligned}
\begin{split}
\label{eq:quad_opt}
    \text{maximize } & \null f(x)=x'Ax \\
    \text{subject to}  & \quad x\in \Delta
\end{split}
\end{aligned}
\end{equation}
\vspace{-0.05cm}

where $x'$ is the transposition of the vector $x$ and $\Delta$ is the standard simplex of $R^n$, defined as $\Delta = \left \{x\in R^n: \sum_{i=1}^n x_i = 1, \text{ and } x_i \geq 0 
\text{ for all } i = 1...n \right\rbrace$. With the assumption of affinity matrix $A$ is symmetric, it is shown by \cite{pavan2007dominant} that if $S$ is an dominant set, then its \textit{weighted characteristic vector} $x^S \in \Delta$ defined as in Eq. \ref{eq:wcv} is the strict local solution of the Standard Quadratic Program in Eq. \ref{eq:quad_opt}. 
\begin{equation}
\begin{aligned}
\label{eq:wcv}
    x_i^S = 
    \begin{cases}
        \frac{w_S(i)}{\sum_{j\in S}w_S(j)},\& \quad i \in S \\
        0,\& \quad \text{otherwise}
    \end{cases}
\end{aligned}
\end{equation}

Conversely, if $x^*$ is a strict local solution to Eq. \ref{eq:quad_opt}, then its \textit{support} $\sigma (x^*)=\left \{ i \in V : x_i >0 \right \} $ is a dominant set of $A$. Thus, a dominant set can be found by localizing a solution of Eq. \ref{eq:quad_opt} by a continuous optimization technique and gathering the support set of the found solution. Notice that the value of a component in the found $x^S \in \Delta$ provides a measure of how strong that component contributes to the cohesiveness of the cluster.

\paragraph{Constrained Dominant Set Framework.} In \cite{zemene2016interactive,mequanint2018dominant} the notion of a {constrained dominant set} is introduced, which aims at finding a dominant set constrained to contain vertices from a given seed set $S \subseteq V$.  
Based on the edge-weighted graph definition with affinity matrix $A$, a parameterized family of quadratic programs is defined as in Eq. \ref{eq:cds_opt} \cite{mequanint2018dominant} for the set $S$ and a parameter $\alpha > 0$,

\vspace{-0.05cm}
\begin{equation}
\begin{aligned}
\label{eq:cds_opt}
    \text{maximize } &\null f_S^\alpha(x)=x'(A-\alpha \hat{I}_S)x \\
    \text{subject to} &\quad x\in \Delta
\end{aligned}
\end{equation}
where $\hat{I}_S$ is the $n \times n$ diagonal matrix whose elements are set to $1$ if the corresponding vertices are in $V\setminus S$ and to $0$ otherwise. 
It is theoretically proven, and empirically illustrated for interactive image segmentation \cite{mequanint2018dominant}, that if $S$ is the set of vertices selected by the user, by setting $\alpha > \lambda_{\text{max}}(A_{V \setminus S})$ it is guaranteed that all local solutions of (\ref{eq:cds_opt}) will have a support that necessarily contains at least one element of $S$. Here, $\lambda_{\text{max}}$ is the largest eigenvalue of the principal submatrix of $A$ indexed by elements of $V\setminus S$. 

In order to find constrained dominant sets by solving the aforementioned quadratic optimization problem (\ref{eq:cds_opt}), \cite{mequanint2018dominant} used Replicator Dynamics that is developed and studied in evolutionary game theory \cite{pavan2007dominant}. In this work we use Infection and Immunization Dynamics (InImDyn) \cite{bulo2011graph} which proved to be a faster and as accurate alternative to it.

\section{Proposed approach} 
\vspace{-0.2cm}
We propose to generate full mask predictions (to be used for supervising a semantic segmentation network) by post-processing weak annotations, i.e. scribble annotations, using CDS. Moreover, we propose to use CDS for multiclass clustering of pixels, i.e. semantic segmentation, while previously CDS has been used only for interactive foreground segmentation \cite{mequanint2018dominant,zemene2016interactive}.  

\subsection{Preprocessing step for CDS} 
\emph{Superpixel generation.} A common approach followed by image segmentation works has been using superpixels as input entities instead of image pixels. A superpixel is a group of pixels with similar colors and using superpixels not only provides reduced computational complexity, but also yields computing features on meaningful regions. Among a variety of techniques, i.e. SLIC, 
Oriented Watershed Transform (OWT), 
we have preferred to use the method developed by Felzenszwalb and Huttenlocher \cite{felzenszwalb2004efficient} similar to \cite{uijlings2013selective} which is a fast and publicly available algorithm. Method of Felzenszwalb and Huttenlocher \cite{felzenszwalb2004efficient} has also been used in another weakly-supervised semantic segmentation framework \cite{lin2016scribblesup} experimenting on the same dataset with us.  Proposed method in \cite{felzenszwalb2004efficient} is a graph-based segmentation scheme where a graph is constructed for an image such that each element to be segmented represents a vertex of the graph and dissimilarity, i.e. color differences, between two vertices constitutes a weighted edge. The vertices (or subgraphs) are started to be merged regarding to a merging criteria given in Eq. \ref{eq:merge}, where $e_{ij}$ is the edge between two subgraphs $C_i$ and $C_j$, $w(e)$ is the weight on edge $e$ and MST($C_x$) be the minimum spanning tree of $C_x$. 
\begin{equation}
\label{eq:merge}    
w(e_{ij}) \leq \min_{x \in \{i,j\}} \left ( \max_{e \in \text{MST}(C_x)} w(e) + \frac{k}{|C_x|} \right)
\end{equation}

Here, $ \frac{k}{|C_x|}$ is a threshold function in which $k$ is decided by the user, i.e. high values of $k$ yield to lower number of (large) segments, and vice-versa. Another parameter given by the user is the smoothing factor (we denote by $\sigma_{\text{FH}}$) of the Gaussian kernel that is used to smooth the image at the preprocessing step.  


\emph{Feature extraction.}
Once the superpixels are generated on the image, a feature vector is computed for each superpixel. In the application of CDS model for interactive image segmentation in \cite{mequanint2018dominant}, median of the color of all pixels in RGB, HSV, and L*a*b* color spaces and Leung-Malik (LM) Filter Bank are concatenated 
in the feature extraction process. Differently from \cite{mequanint2018dominant}, we compute the same feature types with ScribbleSup \cite{lin2016scribblesup}, which has experimented on the same dataset with us, that are color and texture histograms denoted by $h_c(.)$ and $h_t(.)$ in Eq. \ref{eq:7}. More specifically, $h_c(x_i)$ is a histogram computed on the color space using 25 bins and $h_t(x_i)$ is a histogram of gradients at the horizontal and vertical orientations where 10 bins are used for each orientation for the superpixel $x_i$. 

\subsection{Application of CDS for full mask predictions}  
\label{sec:cds_app}
In order to generate full mask predictions using the CDS model, an input image is represented as a graph $G$ where vertices depict the superpixels of the image and edge-weights between vertices reflect the similarity between corresponding superpixels. We use scribbles as the given weak annotations in this work which serve as constraints in the CDS implementation. Previously, CDS has been applied for interactive foreground segmentation \cite{mequanint2018dominant} where dominant set clusters covering a set of given nodes $S$ for a single object class were explored. In this work our problem demand for multiclass clustering of pixels. Hence, here $S_c$ represents the manually selected pixels of the class $c$ where $c \in \{ 1,...,C \}$ and $C$ is the number of classes in the dataset, e.g. $C=21$ for PASCAL VOC 2012. 

Accordingly, for each class of scribbles that exist in a given image, by ignoring the existence of the remaining classes in the image we perform foreground segmentation, i.e. 2-class clustering of image pixels, as in \cite{mequanint2018dominant} by computing its CDS's. Thus, for the class $c$ the union of the extracted dominant sets, i.e. $UDS_c = D_1 \cup D_2 \cup ... D_{L}$ if $L$ dominant sets are extracted which contain the set $S_c$, represents the segmented regions of object in class $c$. We then repeat this process for every class that exist in the image using the corresponding $S_c$ information. If a node, i.e. superpixel, is found in more than one class of $UDS_c$, we assign it to the one having the highest value in its weighted characteristic vector $x^{S_c} \in \Delta$ which is found by solving the quadratic program in Eq. \ref{eq:cds_opt} by InImDyn (see Section \ref{sec:cds}). 

\emph{Computation of the Affinity matrix.} Before computing the CDS clusters, the affinity (or similarity) between superpixels should be computed to construct the matrix $A$ in Eq. \ref{eq:cds_opt}. In \cite{mequanint2018dominant}, dissimilarity measurements are transformed to affinity space by using the Gaussian kernel 
$A_{ij}^\sigma = \mathbb{1}_{i \neq j} \exp \left (\frac{|| f_i-f_j ||^2}{2 \sigma^2} \right )$, where $f_i$ is the feature vector of the superpixel $i$, $\sigma$ is the scale parameter for the Gaussian kernel and $\mathbb{1}_P=1$ if $P$ is true, 0 otherwise. Differently from \cite{mequanint2018dominant}, we use the Gaussian kernel in Eq. \ref{eq:7} where different $\sigma$ values are used for different feature types. The kernel in Eq. \ref{eq:7} is also adopted in \cite{lin2016scribblesup} which experiments on the same dataset and uses the same feature types with us.  
\begin{equation}
    \label{eq:7}
    A_{ij} ^{\sigma_c, \sigma_t}=  \mathbb{1}_{i \neq j} \exp \left ( -\frac{||h_c(x_i)-h_c(x_j)||_2^2}{\sigma_c^2} -\frac{||h_t(x_i)-h_t(x_j)||_2^2}{\sigma_t^2} \right )
\end{equation}

\emph{Using different color spaces.} Quality of generated superpixels effects the performance of the segmentation algorithm directly and a number of segmentation works (examples include but not limited to \cite{uijlings2013selective,aslan2017Color-Spaces}) have emphasized that higher segmentation performances can be obtained by using different color transformations of the input image to deal with different scene and lighting conditions. Motivated by the related literature studies \cite{uijlings2013selective,aslan2017Color-Spaces}, we compute superpixels 
in a variety of color spaces with a range of invariance properties. Specifically, we use five color spaces, that were also used in \cite{uijlings2013selective} for determining high quality object locations by employing segmentation as a selective search strategy, that are \emph{Intensity} (grey-scale image), $Lab$, $rgI$ which denotes $rg$ channels of normalized $RGB$ plus intensity, $HSV$, $H$ that denotes the Hue channel of $HSV$. 
We generate superpixels and compute mask predictions using CDS model for each color space of the input image, then we decide the final label for a pixel based on most frequently occurred class label, i.e. by using the scheme of majority voting. In addition to using different color spaces we also vary the threshold parameter $k$ (in Eq. \ref{eq:merge}) to get benefit from a large set of diversification as recommended in \cite{uijlings2013selective}.

\section{Experiments}
\vspace{-0.2cm}



\emph{{Dataset and evaluation.}} 
We trained the models on the 10582 augmented PASCAL VOC training set \cite{hariharan2011semantic} and evaluated them on the 1449 validation set. We used the scribble annotations published in \cite{lin2016scribblesup}. In what follows accuracy is evaluated using \textit{pixel accuracy} ($\sum_i n_{ii}/\sum_i t_i$), \textit{mean accuracy} ($(1/n_{cl})\sum_i n_{ii}/\sum_i t_i$) and \textit{mean Intersection over Union} ($(1/n_{cl})\sum_i n_{ii}/(t_i + \sum_j n_{ji} - n_{ii}$) as in \cite{long2015fully}, where $n_{ij}$ is the number of pixels of class $i$ predicted to belong to class $j$, $n_{cl}$ is the number of different classes, and $t_i =\sum_j n_{ij}$ be the total number of pixels of class $i$. 

\emph{{Implementation details.}} We used the VGG16-based FCN-8s network \cite{long2015fully} of the MatConvNet-FCN toolbox \cite{vedaldi15matconvnet} which we initialized by ImageNet pretrained model, i.e. VGG-VD-16 in \cite{vedaldi15matconvnet}. We trained by SGD with momentum and, similar to \cite{long2015fully}, we used momentum 0.9, weight decay of $5^{-4}$, mini batch size of 20 images and learning rate of $10^{-3}$. With these selected hyperparameters we observed that the pixel accuracy is being converged on the validation set.

Performance of CDS is sensitive to the selection of the $\sigma$ parameter of the Gaussian kernel (see Section \ref{sec:cds_app}) and in \cite{mequanint2018dominant} three different results are reported for different selections of $\sigma$: 1) \textit{CDSBestSigma}, where best $\sigma$ is selected separately for every image; 2)\textit{CDSSingleSigma}, by searching in a fixed range, i.e. 0.05  and  0.2; 3)\textit{CDSSelfTuning}, where $\sigma^2$ is replaced by $\sigma_i \times \sigma_j$, where $\sigma_i=mean(KNN(f_i))$, i.e. the mean of the K-NearestNeighbor of the sample $f_i$, $K$ is fixed to 7. 
To decide values of the $\sigma_c$ and $\sigma_t$ parameters (in Eq. \ref{eq:7}) we followed \textit{CDSBestSigma} strategy in \cite{mequanint2018dominant}. Additionally, in the graph structure we cut the edges between vertices correspond to non-adjacent superpixels vertices by setting the corresponding items to zero in the affinity matrix $A$ like has been done in \cite{lin2016scribblesup}, which has provided better segmentation maps. We then min-max normalized the matrix $A$ to be scaled in the range of $[0,1]$ and symmetrized it.  

\emph{{Performance evaluation.}}
\label{sec:perf_col}
We first explored the performance using different color spaces on the predicted full annotations of 10582 images (denoted by \textit{PredSet} to mention \say{Predicted Set} in Table \ref{tab1}), before training the network with them. Then, by training the network with the Predicted Sets we report performance on the Test Set, i.e. PASCAL VOC 2012 Val set. In the implementation of the superpixel generation of \cite{felzenszwalb2004efficient} we used smoothing factor of $\sigma_{\textit{FH}} = 0.8$  (\textit{FH} stands for Felzenszwalb and Huttenloche \cite{felzenszwalb2004efficient}) in the experiments of Table \ref{tab1}. For each color space we performed majority voting (denoted by \textit{MV} both in Table \ref{tab1} and \ref{tab2}) over obtained maps with $k=\{225,250,300,400\}$ (in Eq. \ref{eq:merge}). 

We see at Table \ref{tab1} that using different color spaces affects the quality of the predicted full annotations (\textit{PredSet}) and highest quality mask predictions in terms of mIoU are obtained when we use the Intensity (66.51\%). Performing majority voting over maps obtained in all color spaces provided highest quality mask predictions for both CDS (73.28\%) and GraphCut (63.51\%). We then trained the network with the predicted sets of \textit{CDS-Intensity}, \textit{CDS-MV}, \textit{GraphCut-MV} and published full mask annotations and present their performance on the test set in Table \ref{tab1}. We see that by using CDS-MV in training we outperform GraphCut (which was employed in \cite{lin2016scribblesup}) significantly and we are quiet approaching to the performance of fully-annotated mask training (59.2\% vs. 61.6\%).
\raggedbottom
\begin{table}[t!]
\caption{Quality of obtained mask predictions (\textit{PredSet}) and using them in network training performance on the PASCAL VOC 2012 Val set (\textit{TestSet}) (MV: Majority Voting, $^{(*)}$ implementation of GraphCut in our framework.)}
\label{tab1}
\scriptsize
\begin{tabular}{l|l|l|l}
\hline
Color space & mean IoU & Pixel Acc. & mean Acc.\\
\hline
\textit{PredSet}-CDS-$Intensity$ & 66.51 & 89.05 & 75.95\\
\textit{PredSet}-CDS-$Lab$ &  65.47 & 88.36 & 76.15 \\
\textit{PredSet}-CDS-$rgI$  &  64.70 & 88.13 & 75.29\\
\textit{PredSet}-CDS-$HSV$ &  66.49 & 89.27 & 74.60 \\
\textit{PredSet}-CDS-$H$ & 57.16 & 85.12 & 68.21 \\
\hline
\textit{PredSet}-CDS-$MV$ & 73.28 & 91.47 & 82.05  \\
\textit{PredSet}-GraphCut$^{(*)}$-$MV$ & 63.51 & 86.48 & 81.83\\
\hline
\textit{TestSet}-CDS-Intensity & 57.41 & 89.01 & 70.56 \\
\textit{TestSet}-CDS-MV & 59.20 & 89.59 & 73.05 \\
\textit{TestSet}-GraphCut$^{(*)}$-$MV$ &  52.25 & 85.80 & 72.43\\ 
\textit{TestSet}-With Full Masks  & 61.60 & 90.27 & 78.95\\
\hline
\end{tabular}
\end{table}
\raggedbottom

\emph{{Comparison with other full-mask prediction methods.}}
There is a large variety of interactive segmentation algorithms that can be used for full mask prediction to train a semantic segmentation network. 
To be as fair as possible we make comparison with the reported performances of the methods that are carried on in similar conditions with us, e.g. the ones which employ scribbles as weak annotations, achieve network training using cross entropy loss computed over all pixel predictions but not only on given weak annotations, and do not iterate between the shallow segmentation method and network training with the obtained mask predictions as in ScribbleSup \cite{lin2016scribblesup}. On the other hand, we performed the Graph Cut algorithm employed in ScribbleSup \cite{lin2016scribblesup} in our framework by using the published code\footnote{\vspace{0.8cm} mouse.cs.uwaterloo.ca/code/gco-v3.0.zip} referred in \cite{lin2016scribblesup} and present its performance. In fact, our approach can be considered as the first iteration step of such an iterative scheme, and it can be extended to be used in further iterations by updating initial scribble annotations by considering network scores obtained with high confidence. 

Considering the above issues we compare with the methods whose accuracy on the test set is reported when their mask predictions are used to train a segmentation network. Specifically, we refer to the performance results of the popular methods GrabCut \cite{rother2004grabcut}, NormalizedCut \cite{tang2018normalized}, and KernelCut \cite{tang2016normalized} reported in \cite{tang2018normalized}. It is mentioned in \cite{tang2016normalized,tang2018normalized} that for each image pixel, RGB (color) and XY (location) features are concatenated to be used in these algorithms. Then, segmentation proposals generated by them are used to train a VGG16-based DeepLab-Msc-largeFOV network \cite{chen2014semantic}. It is reported in \cite{chen2014semantic} that DeepLab-Msc-largeFOV, which employs atrous convolution and multiscale prediction, outperforms FCN-8s by around 9\% (71.6\% vs. 62.2\%) at PASCAL VOC 2012 validation set when trained by full mask annotations, which provides an advantage at comparative works. On the other hand, we also present the performance gap between weak and full mask training to provide a more fair comparison in Table \ref{tab2}. In Table \ref{tab2}, the performance results of full mask training (64.1 \%), GrabCut \cite{rother2004grabcut}, NormalizedCut \cite{tang2018normalized}, and KernelCut \cite{tang2016normalized} are acquired from \cite{tang2018normalized}.  

\begin{table}[t!]
\caption{Performance comparison on PASCAL VOC 2012 val set.}
\label{tab2}
\scriptsize
\begin{tabular}{l|l|l}
\hline
Method & mIoU & Gap between full and weak supervision \\
\hline
With Full Masks \cite{tang2018normalized} & 64.1 & - \\
GrabCut \cite{rother2004grabcut} & 55.5 & 8.6 \\
NormalizedCut \cite{tang2018normalized} & 58.7 & 5.4 \\
KernelCut  \cite{tang2016normalized} & 59.8 & 4.3 \\
\hline
With Full Masks & 61.6 & - \\ 
GraphCut$^{(*)}$-MV$_{(\sigma_{\text{FH}}=0.8)}$ &  52.25  & 9.35 \\ 
CDS-MV$_{(\sigma_{\text{FH}}=0.8)}$ & 59.20 & 2.40 \\
CDS-MV$_{(\sigma_{\text{FHBest}})}$ & 60.22 & 1.38\\
\hline
\end{tabular}
\end{table}

\begin{figure*}[t!]
    \centering
    \includegraphics[width=1\textwidth]{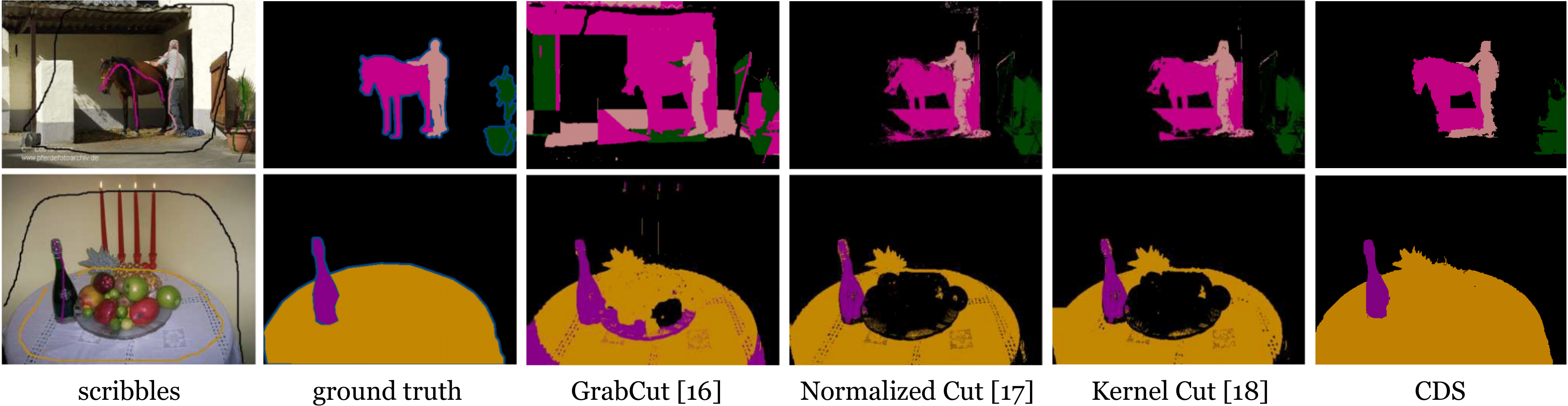}
    \caption{Generated mask predictions (Images for GrabCut \cite{rother2004grabcut}, Normalized Cut \cite{tang2018normalized}, and KernelCut \cite{tang2016normalized} are acquired from \cite{tang2018normalized})}
    \label{fig:1}
\end{figure*}

\begin{figure*}[t!]
    \centering
    \includegraphics[width=1\textwidth]{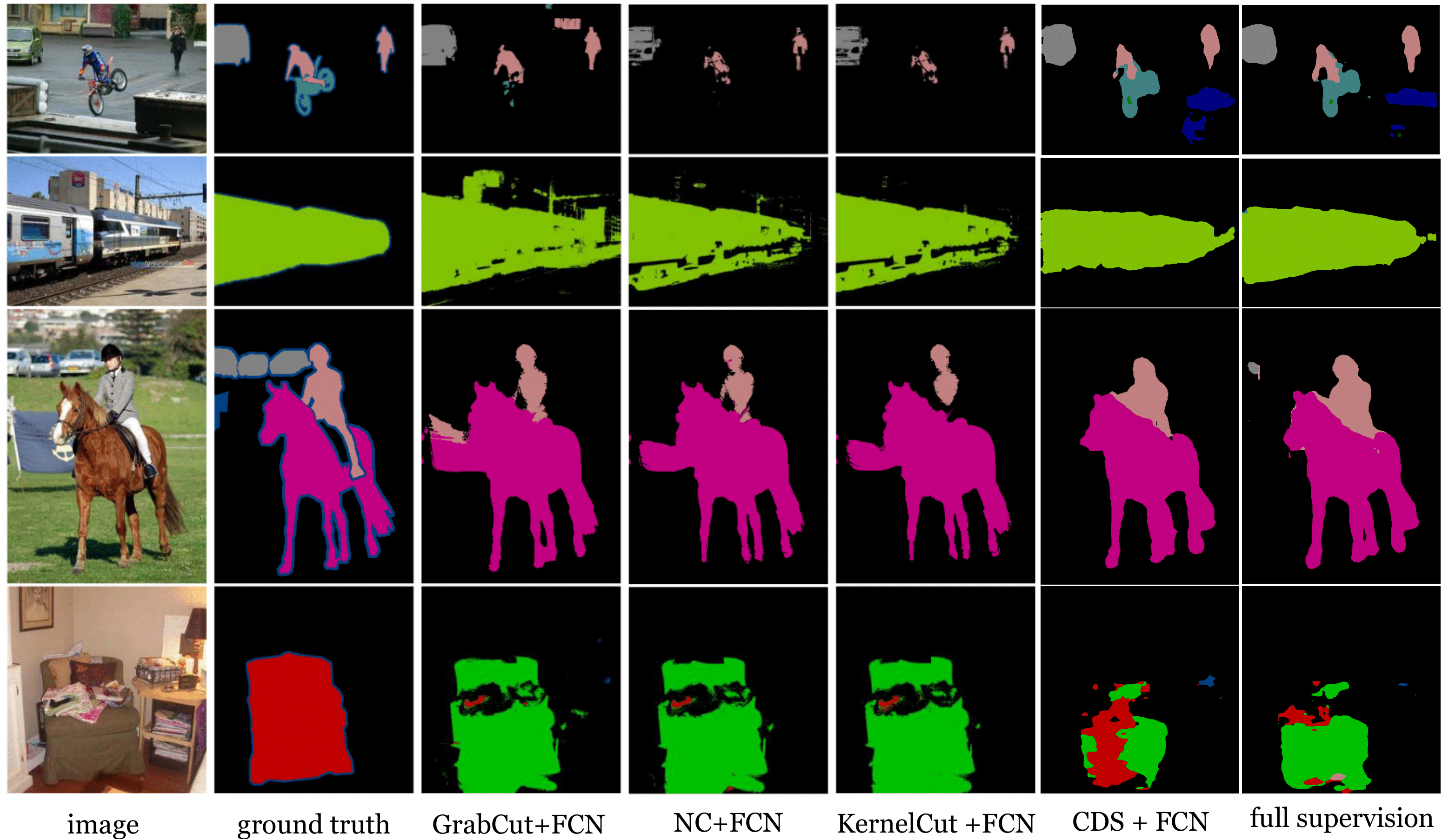}
    \caption{Testing on PASCAL VOC 2012 \textit{val} set. (Images for GrabCut \cite{rother2004grabcut}, Normalized Cut \cite{tang2018normalized}, and KernelCut \cite{tang2016normalized} are acquired from \cite{tang2018normalized})}
    \label{fig:2}
\end{figure*}

For CDS, we train with mask predictions generated by two different selections of $\sigma_{\text{FH}}$: \textit{(i)} $\sigma_{\text{FH}}=0.8$ (corresponding to \textit{PredSet}-CDS-$MV$ in Table \ref{tab1}); and \textit{(ii)} $\sigma_{\text{FHBest}}$, where
we selected the best among $\sigma_{\text{FH}}=0.7$ and $\sigma_{\text{FH}}=0.8$ for each image. It can be seen at the segmentation performances on the \textit{val} set given in Table \ref{tab2} that we outperform the literature works at $\sigma_{\text{FHBest}}$ (60.22\%), and we are superior at both parameter selections in terms of performance gap between full and weak supervision, i.e. we approach to the performance of our full mask training (61.6\%) by 2.4\% and 1.38\% at selection of $\sigma_{\text{FH}}=0.8$ and $\sigma_{\text{FHBest}}$, respectively. Two example images from the generated set, i.e. PredSet, of $\sigma_{\text{FHBest}}$ are presented in Figure \ref{fig:1}. Figure \ref{fig:2} shows examples from testing on the \textit{val} set when it is trained by \textit{PredSet-CDS-MV}$_{\sigma_{\text{FHBest}}}$. It can be seen in Figure \ref{fig:1} and \ref{fig:2} that our results are the ones most closest to the ground truth of input images. 

\vspace{-0.4cm}
\section{Conclusions}
\vspace{-0.4cm}
In this paper we have proposed to apply Constrained Dominant Set (CDS) model, which is proved to be an effective method compared to state-of-the-art interactive segmentation algorithms, for propagating weak scribble annotations of a given set of images to obtain the multi-labeled full mask predictions of them. Achieved mask predictions are then used to train a Fully Convolutional Network for semantic segmentation. While CDS has been applied for pixelwise binary classification problem, it has not been explored for semantic segmentation before and this paper presents our work in this direction. Experimental results showed that proposed approach generates higher quality full mask predictions than the existing methods that have been adopted for weakly-supervised semantic segmentation in literature works.
\vspace{-0.4cm}
\bibliographystyle{splncs04}
\bibliography{WSSS_CDS}
\end{document}